\documentclass{article}

\usepackage{microtype}
\usepackage{graphicx}
\usepackage{subfigure}
\usepackage{booktabs} % for professional tables

\usepackage{hyperref}
\usepackage[utf8]{inputenc} % allow utf-8 input
\usepackage[T1]{fontenc}    % use 8-bit T1 fonts
\usepackage{url}            % simple URL typesetting
\usepackage{booktabs}       % professional-quality tables
\usepackage{amsfonts}       % blackboard math symbols
\usepackage{nicefrac}       % compact symbols for 1/2, etc.
\usepackage{microtype}      % microtypography
\usepackage{chapterbib}
\usepackage{amsmath,amsfonts,bm}

\def\eqref#1{equation~\ref{#1}}
\def\1{\bm{1}}

\DeclareMathAlphabet{\mathsfit}{\encodingdefault}{\sfdefault}{m}{sl}
\SetMathAlphabet{\mathsfit}{bold}{\encodingdefault}{\sfdefault}{bx}{n}

\usepackage{enumitem}
\usepackage{url}
\usepackage{amsmath}
\usepackage{graphicx}
\usepackage{multirow}
\usepackage{amsthm}

\newcommand{\data}{\mathcal{D}}

\usepackage[accepted]{icml2020}

\icmltitlerunning{Hallucinative Topological Memory for Zero-Shot Visual Planning}

\begin{document}

\twocolumn[
\icmltitle{Hallucinative Topological Memory for Zero-Shot Visual Planning}

% It is OKAY to include author information, even for blind
% submissions: the style file will automatically remove it for you
% unless you've provided the [accepted] option to the icml2020
% package.

% List of affiliations: The first argument should be a (short)
% identifier you will use later to specify author affiliations
% Academic affiliations should list Department, University, City, Region, Country
% Industry affiliations should list Company, City, Region, Country

% You can specify symbols, otherwise they are numbered in order.
% Ideally, you should not use this facility. Affiliations will be numbered
% in order of appearance and this is the preferred way.
\icmlsetsymbol{equal}{*}

\begin{icmlauthorlist}
\icmlauthor{Kara Liu}{equal,to}
\icmlauthor{Thanard Kurutach}{equal,to}
\icmlauthor{Christine Tung}{to}
\icmlauthor{Pieter Abbeel}{to}
\icmlauthor{Aviv Tamar}{to,goo}
% \icmlauthor{Tateu H.~Yasehe}{ed,to,goo}
% \icmlauthor{Aaoeu Iasoh}{goo}
% \icmlauthor{Buiui Eueu}{ed}
% \icmlauthor{Aeuia Zzzz}{ed}
% \icmlauthor{Bieea C.~Yyyy}{to,goo}
% \icmlauthor{Teoau Xxxx}{ed}
% \icmlauthor{Eee Pppp}{ed}
\end{icmlauthorlist}

\icmlaffiliation{to}{Berkeley AI Research, University of California, Berkeley}
\icmlaffiliation{goo}{Technion}
% \icmlaffiliation{ed}{School of Computation, University of Edenborrow, Edenborrow, United Kingdom}

\icmlcorrespondingauthor{Kara Liu}{karamarieliu@berkeley.edu}
\icmlcorrespondingauthor{Thanard Kurutach}{thanard.kurutach@berkeley.edu}

% You may provide any keywords that you
% find helpful for describing your paper; these are used to populate
% the "keywords" metadata in the PDF but will not be shown in the document
\icmlkeywords{Machine Learning, ICML}

\vskip 0.3in
]

% this must go after the closing bracket ] following \twocolumn[ ...

% This command actually creates the footnote in the first column
% listing the affiliations and the copyright notice.
% The command takes one argument, which is text to display at the start of the footnote.
% The \icmlEqualContribution command is standard text for equal contribution.
% Remove it (just {}) if you do not need this facility.

%\printAffiliationsAndNotice{}  % leave blank if no need to mention equal contribution
\printAffiliationsAndNotice{\icmlEqualContribution} % otherwise use the standard text.

\begin{abstract}
In visual planning (VP), an agent learns to plan goal-directed behavior from observations of a dynamical system obtained offline, e.g., images obtained from self-supervised robot interaction. Most previous works on VP approached the problem by planning in a learned latent space, resulting in low-quality visual plans, and difficult training algorithms. Here, instead, we propose a simple VP method that plans directly in image space and displays competitive performance. 
We build on the semi-parametric topological memory (SPTM) method: image samples are treated as nodes in a graph, the graph connectivity is learned from image sequence data, and planning can be performed using conventional graph search methods. We propose two modifications on SPTM. First, we train an energy-based graph connectivity function using contrastive predictive coding that admits stable training. Second, to allow zero-shot planning in new domains, we learn a conditional VAE model that generates images given a context of the domain, and use these \emph{hallucinated} samples for building the connectivity graph and planning. We show that this simple approach significantly outperform the SOTA VP methods, in terms of both plan interpretability and success rate when using the plan to guide a trajectory-following controller. Interestingly, our method can pick up non-trivial visual properties of objects, such as their geometry, and account for it in the plans. 
\end{abstract}

% Two or three meaningful keywords should be added here
% \keywords{Visual Planning, Model-based RL, Representation Learning} 

%===============================================================================
\vspace{-0.5em}
\section{Introduction}
\vspace{-0.5em}

We are interested in goal-directed planning problems where the state observations are high-dimensional images, the system dynamics are not known, and only a data set of state transitions is available. In particular, given a starting state observation and a goal state observation, we wish to generate a sequence of actions that transition the system from start to goal. One application for such problems is in self-supervised robot learning, where it is relatively easy to acquire such data by letting the robot explore its environment randomly, and the problem becomes how to process this data for solving various tasks \cite{nair2017combining, wang2019learning, pinto2016supersizing, finn2017deep}.

Given a reward function, deep reinforcement learning (RL) can plan with high-dimensional inputs, and batch off-policy RL algorithms \cite{lange2012batch} can be applied to the problem above~\citep{haarnoja2018soft, ebert2018visual, mnih2015human, schulman2015trust}. However, goal-based planning is a sparse reward task, which is known to be difficult for RL \cite{andrychowicz2017hindsight}. Moreover, RL provides black-box decision policies, which are not interpretable, and can only be evaluated by running them on a robot.
Addressing both data-driven modeling and interpretability, the \emph{visual planning} (VP) paradigm seeks to first generate a visual plan -- a sequence of images that transition the system from start to goal, which can be understood by a human observer -- and only then take actions that follow the plan using visual servoing methods.

Bearing similarity to model-based RL \cite{sutton1998introduction}, most VP approaches learn a low-dimensional latent variable model for the system dynamics, and plan a state-to-goal sequence by searching in the latent space \cite{kurutach2018learning, asai2019unsupervised, ebert2018visual, hafner2018learning, nair2019hierarchical}. There are two shortcomings to this approach: training deep generative models with a structured latent space can be tricky in practice \cite{watter2015embed, kurutach2018learning}, and consequentially, the resulting visual plans are often of low visual fidelity.

In this work, we propose a simple VP method that plans directly in image space. We build on the semi-parametric topological memory (SPTM) method proposed by Savinov et al. \citep{savinov2018semi}. In SPTM, images collected offline are treated as nodes in a graph and represent the possible states of the system. To connect nodes in this graph, an image classifier is trained to predict whether pairs of images were `close' in the data or not, effectively learning which image transitions are feasible in a small number of steps. The SPTM graph can then be used to generate a visual plan -- a sequence of images between a pair of start and goal images -- by directly searching the graph. SPTM has several advantages, such as producing highly interpretable visual plans and the ability to plan long-horizon behavior. Here, we ask -- is such a simple scheme competitive with VP methods that plan in latent space?

\begin{figure*}[h!]
\vspace{-0.5em}
\centering
\includegraphics[width=\textwidth]{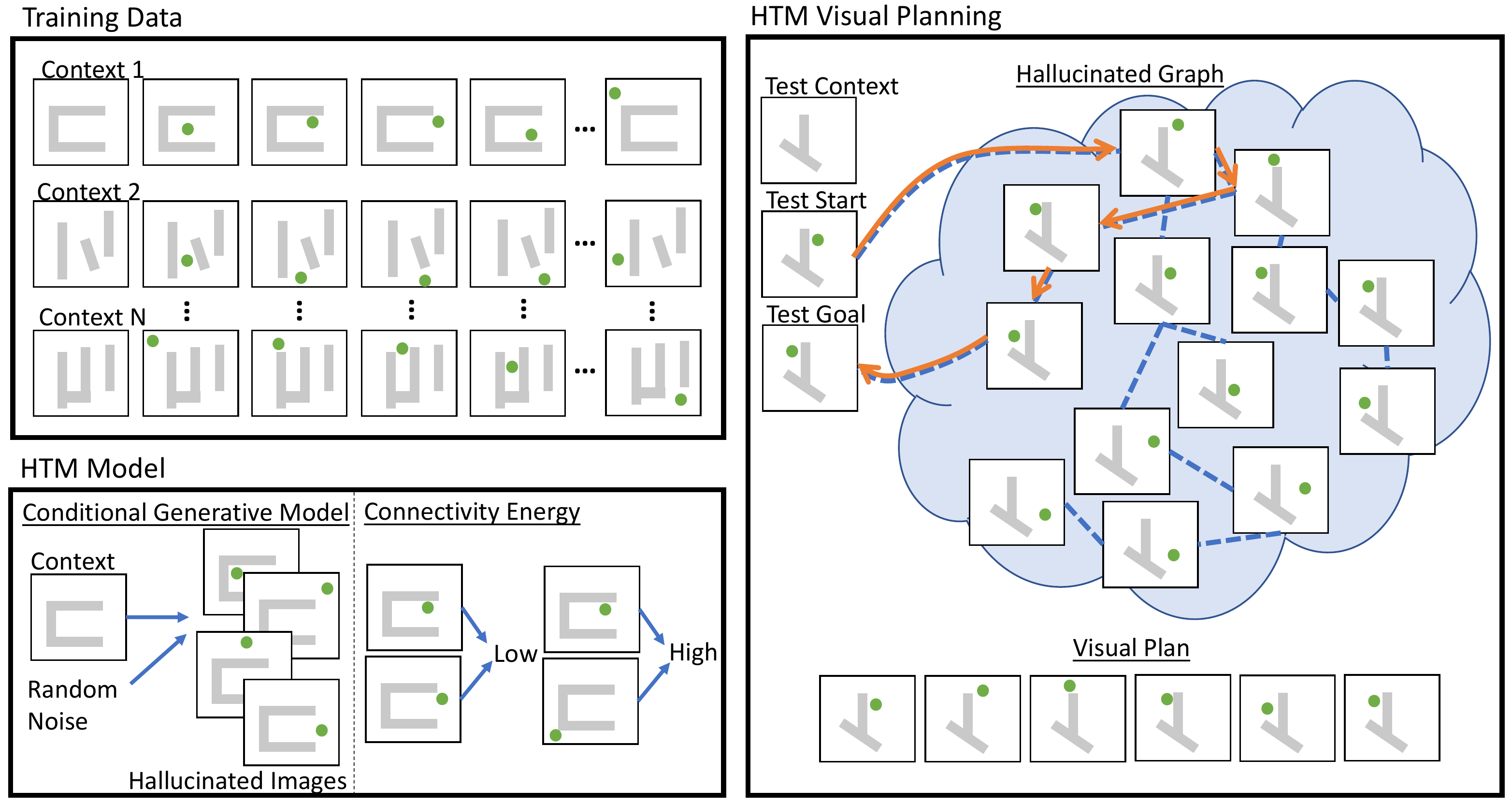}
\vspace{-0.5em}
\caption{HTM illustration. \textbf{Top left:} data collection. In this illustration, the task is to move a green object between gray obstacles. Data consists of multiple obstacle configurations (contexts), and images of random movement of the object in each configuration. \textbf{Bottom left:} the elements of HTM. A conditional generative model is trained to hallucinate images of the object and obstacles conditioned on the obstacle image context. A connectivity energy model is trained to score pairs of images based on the feasibility of their transition. \textbf{Right:} HTM visual planning. Given a new context image and a pair of start and goal images, we first use the conditional generative model to hallucinate possible images of the object and obstacles. Then, a connectivity graph (blue dotted lines) is computed based on the connectivity energy, and we plan for the shortest path from start to goal on this graph (orange solid line). For  plan execution, visual servoing is later used to track the image sequence.}
\label{fig:htm_data}
\end{figure*}

To answer this question, we need to address a limitation of SPTM compared to VP methods such as visual foresight \cite{finn2017deep, ebert2018visual}. Since SPTM builds the visual plan directly from images in the data, when the environment changes -- for example, the lighting varies, the camera is slightly moved, or other objects are displaced -- SPTM requires \emph{recollecting} images in the new environment; in this sense, SPTM \emph{does not generalize in a zero-shot sense}. To tackle this issue, we assume that the environment is described using some context vector, which can be an image of the domain or any other observation data that contains enough information to extract a plan (see Figure~\ref{fig:htm_data} top left). 
We then train a conditional generative model that hallucinates possible states of the domain conditioned on the context vector. Thus, given an unseen context, the generative model hallucinates exploration data without requiring actual exploration. Using the hallucinated images, we can then perform planning in image space. 

Additionally, similar to \citep{eysenbach2019search}, we find that training the graph connectivity classifier as originally proposed by \citep{savinov2018semi} requires extensive manual tuning. We replace the vanilla classifier used in SPTM with an energy-based model that employs a contrastive loss. We show that this alteration drastically improves planning robustness and quality. Finally, for planning, instead of connecting nodes in the graph according to an arbitrary threshold of the connectivity classifier, as in SPTM, we cast the planning as an inference problem, and efficiently search for the shortest path in a graph with weights proportional to the inverse of a proximity score from our energy model.
Empirically, we demonstrate that this provides much smoother plans and barely requires any hyperparameter tuning. 
We term our approach Hallucinative Topological Memory (HTM). A visual overview of our algorithm is presented in Figure \ref{fig:htm_data}.

\vspace{-0.5em}

We evaluate our method on a set of simulated VP problems that require non-myopic planning, and accounting for non-trivial object properties, such as geometry, in the plans. In contrast with prior work, which only focused on the success of the method in executing a task, here we also measure the \emph{interpretability} of visual planning, through mean opinion scores of features such as image fidelity and feasibility of the image sequence.
In both measures, HTM outperforms state-of-the-art data-driven approaches such as visual foresight \citep{ebert2018visual} and the original SPTM. The codebase is released at \url{https://github.com/thanard/hallucinative-topological-memory}.

%===============================================================================
\vspace{-0.5em}
\section{Preliminaries}
\vspace{-0.5em}

\textbf{Context-Conditional Visual Planning and Acting (VPA) Problem.} 
We consider the context-conditional visual planning problem from \citep{kurutach2018learning,wang2019learning}. Consider deterministic and fully-observable environments $\mathcal{E}_1, ..., \mathcal{E}_N$ that are sampled from an environment distribution $P_\mathcal{E}$. Each environment $\mathcal{E}_i$ can be described by a context vector $c_i$ that entirely defines the dynamics $o_{t+1}^i = m(o_t^i, a_t^i| c_i)$, where $o_t^i, a_t^i$ are the observations and actions, respectively, at timestep $t$ from context $c_i$. For example, in the illustration in Figure~\ref{fig:htm_data}, the context could represent an image of the obstacle positions, which is enough to predict the possible movement of objects in the domain.\footnote{We used such a context image in our experiments. We assume that in a practical application, observing the domain without the robot would be feasible, making this setting relevant to applications.}
As is typical in VP problems, we assume our data $\data = \{o_1^i, a_1^i,...,o_{T_i}^i, c_i\}_{i\in\{1,...,N\}}$ is collected in a self-supervised manner, and that in each environment $\mathcal{E}_i$, the observation distribution is defined as $P_o(\cdot|c_i)$. 

At test time, we are presented with a new environment, its corresponding context vector $c$, and a pair of start and goal observations $o_{start}, o_{goal}$.
Our goal is to use the training data to build a planner  $\mathcal{K}_h(o_{start}, o_{goal}, c)$ and an h-horizon policy $\pi_h(o_{current},o_{target})$. 
The planner's task is to generate a sequence of observations between $o_{start}$ and $o_{goal}$, in which any two consecutive observations are reachable within $h$ time steps. The policy outputs an action that brings the current image to the target image within $h$ steps which can be used to follow the generated plan. This requires both the planner and the policy to work together in zero-shot.

In this work, we first evaluate the planner and policy separately -- the planner by measuring the fidelity of its plans, and the policy, by measuring its success rate in tracking a feasible plan. We then also evaluate the combined planner+policy by measuring the total success rate of the policy applied to the planned trajectories. For simplicity we will omit the subscript $h$ for the planner and the policy.

\textbf{Semi-Parametric Topological Memory (SPTM)} \citep{savinov2018semi} is a visual planning method that can be used to solve a special case of VPA, 
where there is only a single training environment, $\mathcal{E}$ and no context image. SPTM builds a memory-based planner and an inverse-model controller. At training, a classifier $R$ is trained to map two observation images $o_i,o_j$ to a score $\in [0,1]$ representing the feasibility of the transition, where images that are $ \leq h$ steps apart are labeled positive and images that are $\geq l$ are negative. The policy is trained as an inverse model $L$, mapping a pair of observation images $o_i,o_j$ to an appropriate action $a$ that transitions the system from $o_i$ to $o_j$.

Given an unseen environment $\mathcal{E^*}$, new observations are \textit{manually} collected and organized as nodes in a graph $G$. Edges in the graph connect observations $o_{i}, o_{j}$ if $R(o_{i}, o_{j}) \geq s_{shortcut}$, where $s_{shortcut}$ is a manually defined threshold. To plan, given start and goal observations $o_{start}$ and $ o_{goal}$, SPTM first uses $R$ to localize, i.e., find the closest nodes in $G$  to ${o}_{start}$ and ${o}_{goal}$. A path is found by running Dijkstra's algorithm, and the method then selects a waypoint $o_{w_i}$ on the path which represents the farthest observation that is still feasible under $R$. Since both the current localized state $o_{i}$ and its waypoint $o_{w_i}$ are in the observation space, we can directly apply the inverse model and take the action $a_i$ where $a_i = L(o_i, o_{w_i})$. After localizing to the new observation state reached by $a_i$, SPTM repeats the process until the node closest to $ {o}_{goal}$ is reached. 

\textbf{Contrastive Predictive Coding (CPC)} \citep{oord2018representation} is a method for learning low-dimensional representations of high-dimensional sequential data. CPC learns both an encoding of the data at every time step, and an energy function for any two observations in different time steps in suggesting their temporal correlation. A non-linear encoder $g_{enc}$ encodes the observation $o_t$ to a latent representation $z_t = g_{enc}(o_t)$. 
We maximize the mutual information between the latent representation $z_t$ and future observation $o_{t+k}$ with a log-bilinear model\footnote{The original CPC model has an additional autoregressive memory variable~\cite{oord2018representation}. We drop it in our formulation as our domains are fully observable and do not require memory.
}
$f_k(o_{t+k}, o_t) = \exp (z_{t+k}^T W_k z_t).$ This model is trained to be proportional to the density ratio $p(o_{t+k}|z_t)/p(o_{t+k})$ by the CPC loss function: the cross entropy loss of correctly classifying a positive sample from a set of random observations consisting of $1$ positive sample $(o_t, o_{t+k})$ from the paired data $\data_{pair}$ and $N - 1$ negative samples $(o_t, o')$ where $o'$ is sampled separately from the full data $\data_{single}$:
\vspace{-2em}
\[\mathcal{L}_{CPC} = -\mathbb{E}_{(o_t, o_{t+k})\sim \data_{pair}}\left[\log\frac{f_k(o_{t+k}, o_t)}{\sum_{\tilde{o} \in \data_{single}}f_k(\tilde{o}, o_t)}\right].\]

Note that the model $f_k$ is not necessary symmetric, and therefore can capture asymmetric transition in the data. $f_k$ can also be viewed as an inverse energy model whose outputs are high for positive samples and low for negative samples.

%===============================================================================

\vspace{-0.5em}
\section{Hallucinative Topological Memory}
\vspace{-0.5em}
By planning directly in image space, and composing the plan from real images (vs. planning in a learned latent space \cite{kurutach2018learning}), SPTM is guaranteed to produce high-fidelity visual plans. In addition, SPTM has been shown to solve long-horizon planning problems such as navigation from first-person view~\cite{savinov2018semi}. However, \emph{SPTM is not zero-shot}: even a small change to the training environment requires collecting substantial exploration data for building the planning graph. This can be a limitation in practice, especially in robotic domains, as any interaction with the environment requires robot time, and exploring a new environment can be challenging (indeed, \citealt{savinov2018semi} applied manual exploration). 
In addition, similarly to \citet{eysenbach2019search}, we found that training the connectivity classifier as proposed by \citet{savinov2018semi} requires extensive hyperparameter tuning.

In this section, we propose an extension of SPTM to overcome these two challenges by employing three ideas -- (1) using a conditional generative model such as CVAE \citep{sohn2015learning} or CGAN \citep{mirza2014conditional} to hallucinate samples in a zero-shot setting, (2) using contrastive loss for a more robust score function and planner, and (3) planning based on an approximate maximum likelihood formulation of the shortest path under uniform state distribution. We call this approach Hallucinative Topological Memory (HTM), and next detail each component in our method.

\subsection{Hallucinating Samples}
\vspace{-0.5em}

We propose a zero-shot learning solution for automatically building the planning graph using only a context vector of the new environment. Our idea is that, after seeing many different environments and corresponding states of the system during training, given a new environment we should be able to effectively \emph{hallucinate} possible system states. We can then use these hallucinations in lieu of real samples from the system in order to build the planning graph. To generate images conditioned on a context, we implement a conditional generative model as depicted in Figure \ref{fig:htm_data}. During training, 
we learn the conditional distribution $p_\theta(o|c).$ During testing, when prompted with a new context vector $c_{test}$, we generate samples $\hat{o}_1, ..., \hat{o}_N \sim p_\theta(o|c_{test})$ in replacement of exploration data.

\subsection{Algorithm} 
\vspace{-0.5em}
We now describe the HTM algorithm. Given  a start observation $o_{start}$, a goal observation $o_{goal}$ sampled from a potentially new environment $\mathcal{E^*}$, and the context vector $c$, we propose a 4-step planning algorithm. 
\vspace{-0.5em}
\begin{enumerate}[noitemsep]
    \item We hallucinate exploration data $\hat{o}_1, ..., \hat{o}_N$ by sampling from the conditional generative model $p_\theta( \cdot | c)$.
    \item We build a fully-connected weighted graph $G(V,E)$ by forming connections between all generated image pairs $\hat{o}_i, \hat{o}_j$ with learned directed edge weight $w_{ij}$. 
    \item We find the shortest path using Dijkstra's algorithm on the learned connectivity graph $G$ between the start and goal node.
    \item We apply a local policy to follow the visual plan, attempting the next node in our shortest path for $h$ time steps, and replan every fixed number of steps until we reach $o_{goal}$. 
\end{enumerate}
\vspace{-0.5em}
In step 2, the weights should reflect difficulty in transitioning from one state to another using a self-supervised exploration policy. The learned connectivity graph $G$ can be viewed as a topological memory upon which we can use conventional graph planning methods to efficiently perform visual planning.
In step 4, for the policy, we train an inverse model which predicts an action given the current observation and a nearby goal observation. In practice, given a transition $(o_t, a_t, o_{t+1})$, we train a deep convolutional neural network $\pi(o_t, o_{t+1}) = \hat{a}_t$ to minimize the L2 loss between $a_t$ and $\hat{a}_t$ \cite{nair2017combining, wang2019learning}. 

\subsection{Learning the Connectivity via Contrastive Loss}
\vspace{-0.5em}
\label{sec:cpc_connect}
A critical component in the SPTM method is the connectivity classifier that decides which image transitions are feasible.
False positives may result in impossible short-cuts in the graph, while false negatives can make the plan unnecessarily long. 
In \cite{savinov2018semi}, the classifier was trained discriminatively, using observations in the data that were reached within $h$ steps as positive examples, and more than $l$ steps as negative examples, where $h$ and $l$ are chosen arbitrarily. In practice, this leads to three important problems. First, this method is known to be sensitive to the choice of positive and negative labeling~\cite{eysenbach2019search}. Second, training data are required to be long, non-cyclic trajectories for a high likelihood of sampling `true' negative samples. However,  self-supervised interaction data often resembles random walks that repeatedly visit a similar state, leading to inconsistent estimates on what constitutes negative data. Third, since the classifier is only trained to predict positively for temporally nearby images and negatively for temporally far away images, its predictions of \textit{medium-distance} images can be arbitrary. This creates both false positives and false negatives, thereby increasing shortcuts and missing edges in the graph.

To solve these problems, we propose to learn a connectivity score using contrastive predictive loss \citep{oord2018representation}. We initialize a CPC encoder $g_{enc}$ that takes in both observation and context, and a density-ratio model $f_k$ that does not depend on the context. Through optimizing the CPC objective, $f_k$ is trained such that positive pairs, which appear sequentially, have higher score, i.e., lower energy, than negative pairs, which are sampled randomly from the data. Thus, it serves as a proxy for the temporal distance between two observations in the sense that sequential observations should have lower energy, leading to a connectivity score for planning in the next section. Compared to the heuristic classification loss in SPTM, the CPC loss is derived from a clear objective: maximize the mutual information between current and future observations. In practice, this results in less hyperparameter tuning and a smoother distance manifold in the representation space mitigating the first and the third problems. 

To tackle the second problem, instead of only sampling negative data within the same trajectory as an anchor image $o_t$ as done in SPTM, we sample any $\Tilde{o}$ that shares the same context $c$ as $o_t$ from the replay buffer. We also find that adding negative data sampled from $p_\theta(.|c)$ can help $f_k$ evaluate more consistently on hallucinated images. Without this trick, we find that the SPTM classifier suffers from false negatives and fails to train on short, cyclical trajectories collected by self-supervised interaction.

\subsection{Edge Weight Selection}
\vspace{-0.5em}
We would like edge weights to reflect the difficulty in transitioning from one state to another according causality in the data -- low weight when the transition is feasible. Based on the connectivity score from the contrastive loss, we proposed two choices of computing the weight $w_{ij}$ from node $j$ to node $i$: (1) an energy model or an inverse of $f_k$, i.e., $1/f_k(i,j)$, and (2) a density ratio or an inverse of normalized $f_k$ over outgoing edges from $j$, i.e., $\sum_{s\in V}f_k(s, j)/f_k(i,j).$ With this heuristic, the shortest path in $G$ tries to predict reachable visual plans. In Appendix, we argue that the shortest path in graph $G$ according to the weights in option 2 leads to maximizing trajectory likelihood bound under uniform data assumption, thus, casting planning as inference.

%===============================================================================

\vspace{-0.5em}
\section{Related work}
\vspace{-0.5em}
\textbf{Reinforcement Learning.} Most of the study of data-driven planning has been under the model-free RL framework~\cite{schulman2015trust, mnih2015human, silver2016mastering}. However, the need to design a reward function, and the fact that the learned policy does not generalize to tasks that are not defined by the specific reward, has motivated the study of model-based approaches. Recently, ~\citet{kaiser2019model, ichter2019robot} investigated model-based RL from pixels on Mujoco and Atari domains, but did not study generalization to a new environment.
\citet{finn2017deep, ebert2018visual} explored model-based RL with image-based goals using visual model predictive control (visual MPC). These methods rely on video prediction, and are limited in the planning horizon due to accumulating errors. In comparison, our method does not predict full trajectories but only individual images, mitigating this problem. Our method is orthogonal to and can be combined with visual MPC as a replacement for the inverse model. 

Concurrently with our work, \citet{nair2019hierarchical} propose a hierarchical visual MPC method that, similarly to our approach, optimizes a sequence of hallucinated images as sub-goals for a visual MPC controller, by searching over the latent space of a CVAE. To compute a search update over a proposed plan, the algorithm evaluates the video prediction score between consecutive subgoals making it more expensive and challenging to optimize as the number of subgoals increases. In practice, it is shown to work with the maximum of 2 subgoals. We also find that it takes approximately to 2 hours to plan a single task comparing to a few seconds in HTM making the algorithm impractical to evaluate without access to a large GPU cluster and especially in a closed-loop setting. 

\textbf{Self-supervised learning.} Several studies investigate planning goal directed behavior from data obtained offline, e.g., by self-supervised robot interaction \cite{agrawal2016learning, pinto2016supersizing}. \citet{nair2017combining} use an inverse model to reach local sub-goals, but require human demonstrations of long-horizon plans. Wang et al. \citep{wang2019learning} solve the visual planning problem using a conditional version of Causal InfoGAN~\citep{kurutach2018learning}. Our work do not tie to specific types of generative models and in the experiments we opted for the CVAE-based approach for stability and robustness.

\textbf{Classical planning and representation learning.}
Studies that bridge between classical planning and representation learning include \citep{kurutach2018learning, asai2018classical, asai2019unsupervised,eysenbach2019search}. These works, however, do not consider zero-shot generalization. While \citet{srinivas2018universal} and \citet{qureshi2019motion} learn representations that allow goal-directed planning to unseen environments, they require expert training trajectories. \citet{ichter2019robot} also generalizes motion planning to new environments, but require a collision checker and valid samples from test environments.

%===============================================================================
\vspace{-0.5em}
\section{Experiments}\label{sec:experiments}
\vspace{-0.5em}

We evaluate HTM on a suite of simulated tasks inspired by robotic manipulation domains. We note that recent work in visual planning (e.g., \citealt{kurutach2018learning,wang2019learning,ebert2018visual}) focused on real robotic tasks with visual input. While impressive, such results can be difficult to reproduce or compare. For example, it is not clear whether manipulating a rope with the PR2 robot \citep{wang2019learning} is more or less difficult than manipulating a rigid object among many visual distractors \citep{ebert2018visual}. Our suite of tasks is reproducible, contains clear evaluation metrics, and our code will be made available for evaluating other algorithms in the future.

\begin{figure}[h]
\centering
\includegraphics[width=\linewidth]{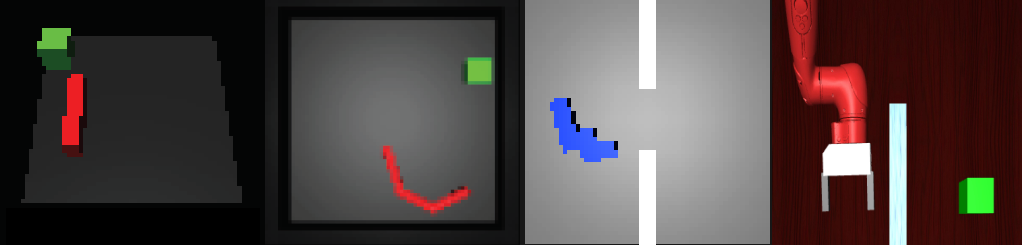}
\vspace{-0.5em}
\caption{ Evaluation suite. Block wall, block wall with complex obstacle, block insertion, and robot manipulation domains.}
\label{fig:samples}
\end{figure}

We consider four domains in varying difficulty using Mujoco simulation~\cite{todorov2012mujoco}, as seen in Figure \ref{fig:samples}:
\vspace{-1em}
\begin{enumerate}[noitemsep]
    \item \textbf{Block wall}: A green block navigates around a static red obstacle, which can vary in position.
    \item \textbf{Block wall with complex obstacle}: Similar to the above, but here the wall is a 3-link object which can vary in position, joint angles, and length, making the task significantly harder. 
    \item \textbf{Block insertion: } Moving a blue block, which can vary in shape, through an opening. 
    \item \textbf{Robot manipulation: } A simulated Sawyer robot reaching and displacing a block. 
\end{enumerate}
 \vspace{-0.5em}
\vspace{-0.5em}

With the first three domains, we aim to assess how well HTM can \textit{generalize} to new environments in a zero-shot manner, by varying the position of the obstacle, the shape of the obstacle, and the shape of the object. 
With the forth domain, we aim to assess whether HTM can plan temporally-extended robotic manipulation.
%study visual planning algorithms by increasing the planning difficulty, which we consider to be the number of steps required to solve the task.

We ask the following questions. First, does HTM improve \textbf{visual plan quality} over state-of-the-art VP methods  \citep{savinov2018semi, ebert2018visual}? Second, how does HTM \textbf{execution success rate} compare to state-of-the-art VP methods? We discuss our evaluation metrics for these attributes in Section \ref{sec:metrics}. 

We compare HTM with two state-of-the-art baselines: 
% To fully assess success of HTM relative to other state-of-the-art VP methods, we run these evaluation metrics on 
SPTM \citep{savinov2018semi} and Visual Foresight \citep{ebert2018visual}. 
% In the first baseline, 
When evaluating zero-shot generalization, SPTM requires samples from the new environment. For a fair comparison, we use the same samples generated by the same CVAE as HTM. Thus, in this case, we only compare between the SPTM classification scores as edge weights in the graph, and our CPC-based scores.\footnote{In practice, we found that exponentiating the SPTM classifier score instead of thresholding worked slightly better, without requiring tuning a threshold. We therefore report results using this method.} 

The same low-level controller is also used to follow the plans. The Visual Foresight baseline trains a video prediction model, and then performs model predictive control (MPC), which searches for an optimal action sequence using random shooting. For the random shooting, we used 3 iterations of the cross-entropy method with 200 sample sequences. The MPC acts for 10 steps and then replans, where the planning horizon $T$ is set to 15 as in the original implementation. Experiments with different horizons yielded worse performance. We use the state-of-the-art video predictor as proposed by Lee et al. \citep{lee2018stochastic} and the public code provided by the authors. 
% \footnote{Code can be located on the website provided in \citep{ebert2018visual} .} 
For evaluating trajectories in random shooting, we studied two cost functions that are suitable for our domains: pixel MSE loss and green pixel distance. The pixel MSE loss computes the pixel distance between the predicted observations and the goal image. This provides a sparse signal when the object pixels in the plan can overlap with those of the goal. We also investigate a cost function that uses prior knowledge about the task -- the position of the moving green block, which is approximated by calculating the center of mass of the green pixels. As opposed to pixel MSE, the green pixel distance provides a smooth cost function which estimates the normalized distance between the estimated block positions of the predicted observations and the goal image. Note that this assumes additional domain knowledge which HTM does not require.

\subsection{Evaluation Metrics} \label{sec:metrics}
\vspace{-0.5em}
We design a set of tests that measure both qualitative and quantitative performance of an algorithm. While the quantitative tests evaluate how successful the algorithm is in solving a task, our qualitative tests provide a measure of \textit{plan interpretability}, which is often desired in practice. 

\textbf{Qualitative evaluation:} Visual plans can be inspected by a human to assess their quality. 
% have the essential property of being intuitive, in that the imagined trajectory is perceptually sensible. 
Since human assessment is subjective, we devised a set of questionnaires, and for each domain, we asked 5 participants to visually score 5 randomly generated plans from each model by answering the following questions: 
(1) \textit{Fidelity}: Does the pixel quality of the images resemble the training data?; (2) \textit{Feasibility}: Is each transition in the generated plan executable by a single action step?; and (3) \textit{Completeness}: Is the goal reachable from the last image in the plan using a single action?
Answers were in the range [0,1], where 0 denotes \textit{No} to the proposed question and 1 means \textit{Yes}. The mean opinion scores are reported for each model.

\textbf{Quantitative} In addition to generating visually sensible trajectories, a planning algorithm must also be able to successfully navigate towards a predefined goal. Thus, for each domain, we selected 20 start and goal images, each with an obstacle configuration unseen during training. Success was measured by the ability to get within some L2 distance to the goal in $n$ steps or less, where the distance threshold and $n$ varied on the domain but was held constant across all models. A controller specified by the algorithm executed actions given an imagined trajectory, and replanning occurred every \textit{r} steps. Specific details can be found in the Appendix.

% HTM plans
\begin{figure}[t]
\centering
\includegraphics[width=\linewidth]{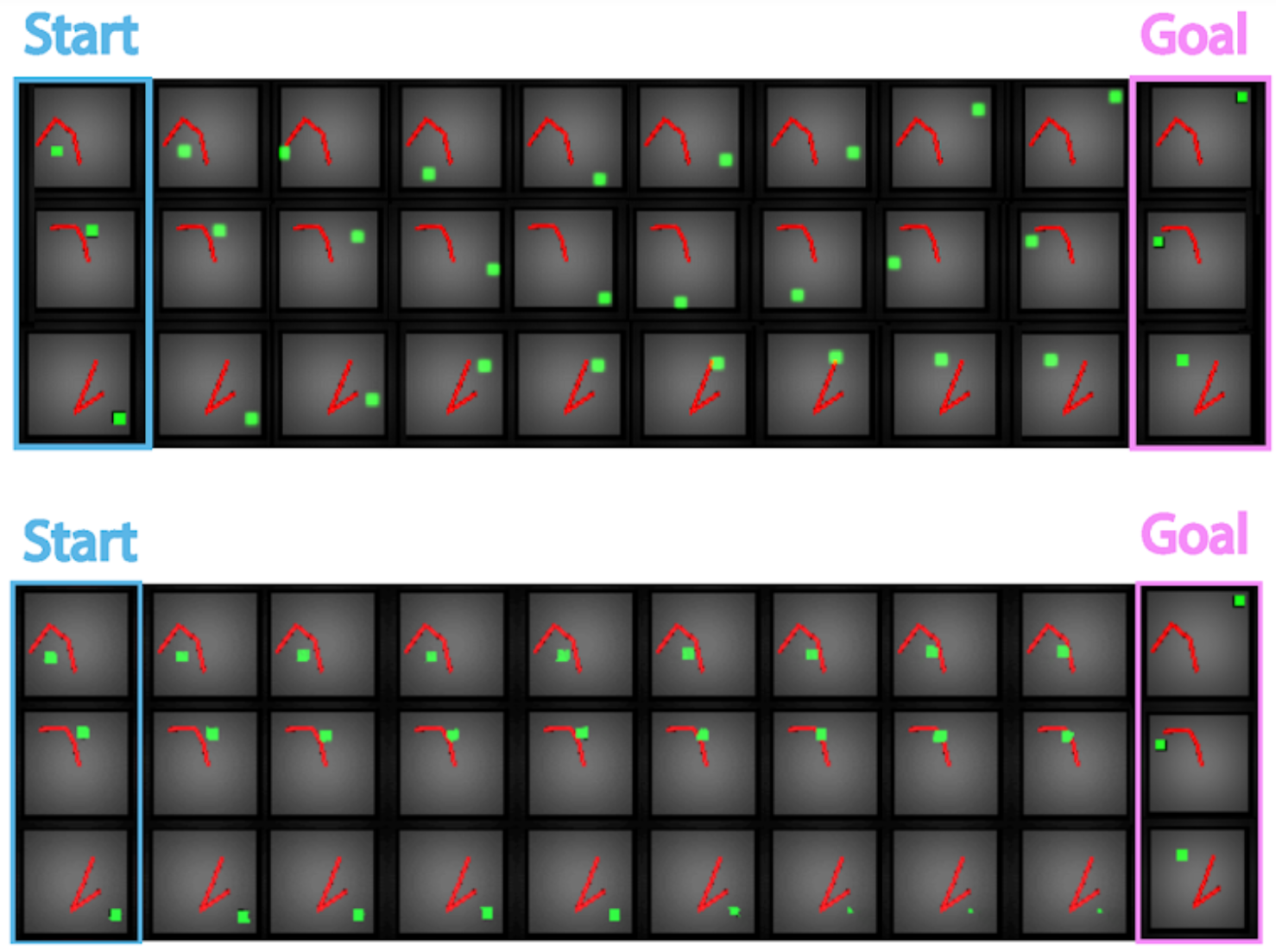}

\caption{HTM plan examples (top 3 rows) and Visual Foresight plan examples (bottom 3 rows). Note Visual Foresight is unable to conduct a long-horizon plan, and thus greedily moves in the direction of the goal state using green pixel distance cost.}

\label{fig:htm_bco}
\vspace{-1em}
\end{figure}

\subsection{Results on Block Domains}
\vspace{-0.5em}
As shown in Table \ref{table:1}, HTM outperforms all baselines in both qualitative and quantitative measurements across the first two domains. 
In the simpler block wall domain, Visual Foresight only succeed with the extra domain knowledge of using the green pixel distance. In the complex obstacle domain, Visual Foresight mostly fails to find feasible plans. SPTM, on the other hand, performed poorly on both tasks, showing the importance of our CPC-based edge weights in the graph.
\textbf{Perhaps the most interesting conclusion from this experiment, however, is that even such visually simple domains, which are simulated, have a single moving object, and do not contain visual distractors or lighting/texture variations, can completely baffle state-of-the-art VP algorithms}. For the complex obstacle domain, we attribute this to the non-trivial geometric information about the obstacle shape that needs to be extracted from the context and accounted for during planning. In comparison, the real-image domains of \citet{ebert2018visual}, which contained many distractors, did not require much information about the shape of the objects for planning a successful pushing action.

\begin{figure}[h!]
\centering
\includegraphics[width=0.40\textwidth]{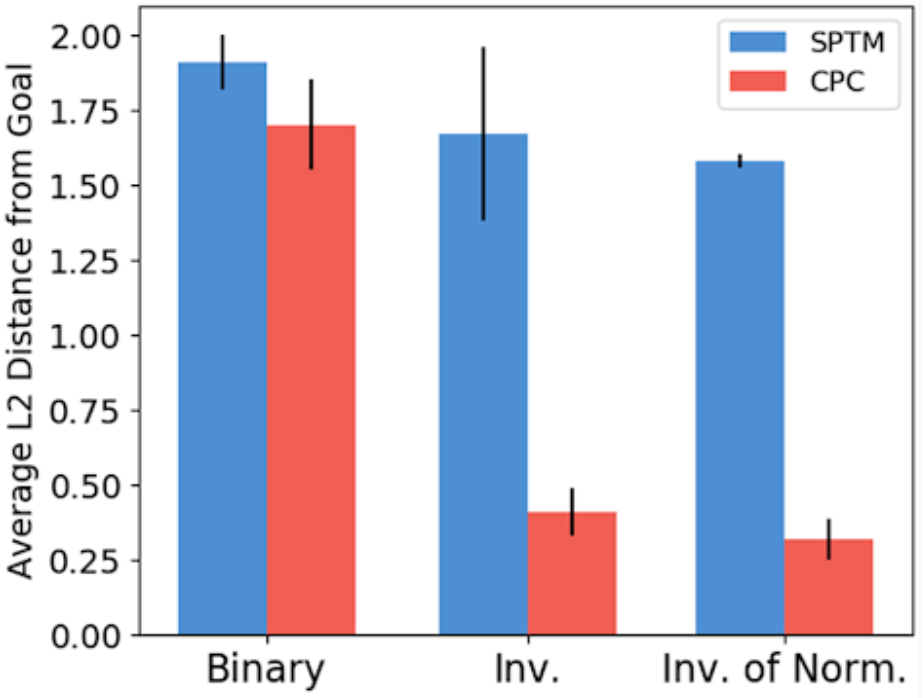}
\caption{Ablation study on weight functions. We show gain in using our proposed score function and weighting function comparing to those proposed in the original SPTM by examining final \textit{average distance} to the goal state for 10 test start/goal pairs on block with complex obstacle domain (the lower the distance, the better). 
For the score function, we denote our proposed energy model structured with contrastive loss as \textit{CPC} and the classifier as proposed in \citep{savinov2018semi} with BCE loss as \textit{SPTM}. 
For the edge weighting function, we test the binary thresholding from the original SPTM paper, our proposed inverse of the score function, and our proposed inverse of the normalized score function.}
\label{fig:ablation}
\end{figure}

In regards to perceptual evaluation, Visual Foresight generates realistic transitions, as seen by the high participant scores for feasibility. However, the algorithm is limited in creating a visual plan within the optimal $T = 15$ timesteps consistent with that of \citep{ebert2018visual}. 
% \footnote{For plans require $> T$ steps, we found that error across the image translations accumulate and the predicted image drastically decreases in interpretability. This optimal value of $T$ is consistent with that of \citep{ebert2018visual}.}  
Thus, when confronted with a challenging task of navigating around a concave shape where the number of timesteps required exceeds $T$, Visual Foresight fails to construct a reliable plan (see Figure \ref{fig:htm_bco}), and thus lacks plan completeness. Conversely, SPTM is able to imagine some trajectory that will reach the goal state. However, as mentioned above and was confirmed in the perceptual scores, SPTM fails to select feasible transitions, such as imagining a trajectory where the block will jump across the wall or split into two blocks. Our approach, on the other hand, received the highest scores of fidelity, feasibility, and completeness. 
Finally, we show in Figure \ref{fig:ablation} the results of our two proposed improvements to SPTM in isolation. The results clearly show that a classifier using contrastive loss outperforms that which uses Binary Cross Entropy (BCE) loss, and furthermore that the inverse of the score function for edge weighting is more successful than the best tuned version of binary edge weights through thresholding -- 0 means no edge connection and 1 means an edge exists.

% Evaluation
\begin{table*}[h!]
\centering
\begin{tabular}{c|c || c|c|c|| c} 
\\
 Algorithms & Domain & Fidelity & Feasibility & Completeness & Execution Success \\
 \hline 
% \multirow{2}{*}{\textbf{HTM (1)}} & 1 & \textbf{0.86 $\pm$ .05} & \textbf{0.84 $\pm$ .16} & \textbf{1.00 $\pm$ .00} & \textbf{100\%} \\
% & 2 & \textbf{0.95 $\pm$ .03} & \textbf{0.92 $\pm$ .11 }& \textbf{1.00 $\pm$ .00} &\textbf{95\%} \\
% \hline
\multirow{2}{*}{\textbf{HTM }}& 1 & \textbf{0.75 $\pm$ .09} & \textbf{0.88 $\pm$ .14} & \textbf{1.00 $\pm$ .00} & \textbf{95\%} \\
 & 2 & \textbf{0.96 $\pm$ .03} & \textbf{0.96 $\pm$ .08 }& \textbf{0.96 $\pm$ .08} &\textbf{100\%} \\
\hline
\multirow{2}{*}{SPTM with CVAE} & 1 & 0.40 $\pm$ .11 & 0.00 $\pm$ .00 & 1.00 $\pm$ .00 & 55\%  \\
& 2 & 0.92 $\pm$ .07 & 0.00 $\pm$ .00 & 1.00 $\pm$ .00 & 30\%\\
\hline
Visual Foresight \cite{ebert2018visual} & 1 & 0.74 $\pm$ .08 & 0.84 $\pm$ .16 & 0.04 $\pm$. 08 & 25\% \\
(pixel MSE loss) & 2 & 0.59 $\pm$ .16 & 0.64 $\pm$ .21 & 0.00 $\pm$ .00 & 0\% \\
\hline
Visual Foresight \cite{ebert2018visual} & 1 & 0.80 $\pm$ .07 & 0.84 $\pm$ .16 & 0.04 $\pm$ .08 & 90\%  \\
(green pixel distance)& 2 & 0.69 $\pm$ .14 & 0.56 $\pm$ .21 & 0.00 $\pm$ .00 & 35\%   \\
\hline
\multirow{2}{*}{Inverse Model} & 1& - & - & - & 20\% \\
&2 & - & - & - & 25\% \\

\end{tabular}
\vspace{-0.5em}
\caption{Qualitative and quantitative evaluation for the the block wall (1) and block wall with complex obstacle (2) domains. Qualitative data also displays the 95\% confidence interval.}
\label{table:1}
\end{table*}
% \vspace{-0.5em}

\begin{table}[h!]
\centering

\begin{tabular}{c|c||c} 
\\
 Algorithms & Difficulty & Execution Success \\
\hline
\multirow{2}{*}{\textbf{HTM}} & \textbf{Easy} & \textbf{100 \%} \\
& \textbf{Hard} & \textbf{70 \%} \\
\hline
\multirow{2}{*}{Visual Foresight  }& Easy & 60 \% \\
 & Hard & 10 \% \\
\hline
\multirow{2}{*}{Inverse Model} & Easy & 90 \% \\
& Hard & 30 \% \\

\end{tabular}
\vspace{-0.5em}
\caption{Quantitative evaluation for block insertion domain. Visual Foresight \cite{ebert2018visual} was trained using pixel MSE loss. }
\end{table}
\vspace{-0.5em}

\begin{figure}[t]
\centering
\includegraphics[width=\linewidth]{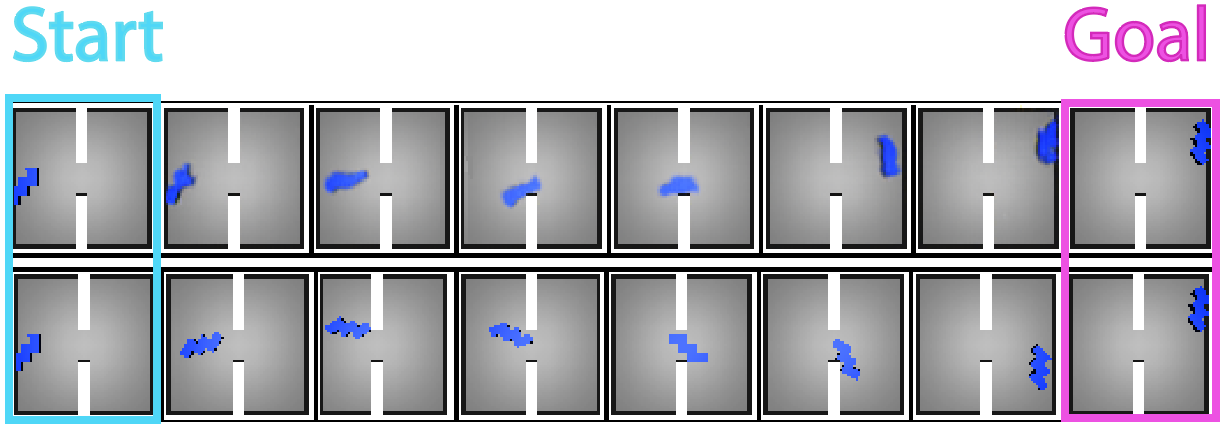}

\caption{HTM plan and execution. The top row demonstrates a generated visual plan on an unseen block configuration, and the bottom displays the execution to follow the plan. }
\label{fig:d3}
\vspace{-1em}

\end{figure}

\begin{figure}[t]
\centering
\includegraphics[width=\linewidth]{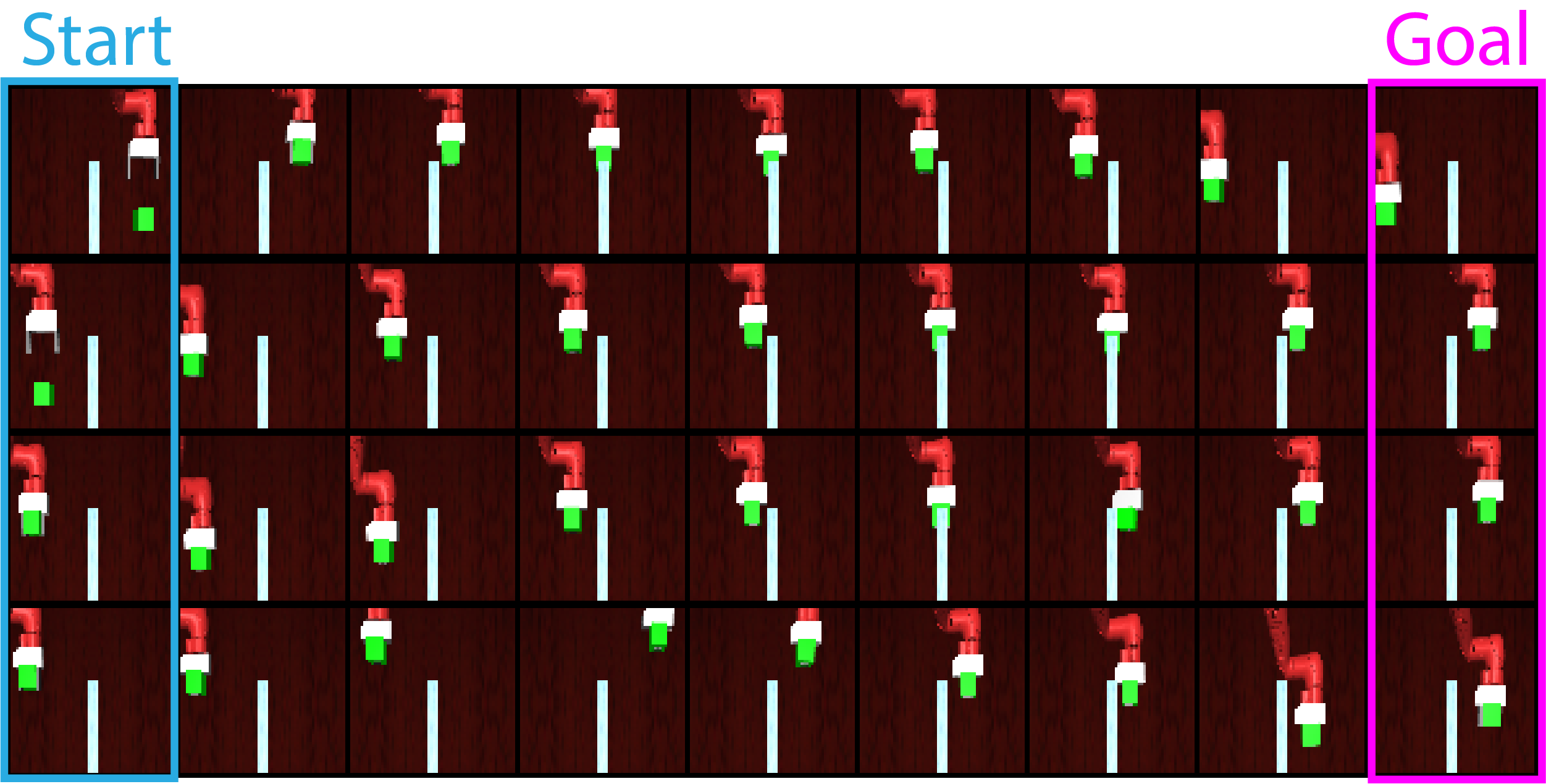}

\caption{HTM plans on the replay buffer. The agent plans to grab the green block and/or go around the obstacle with goal directed planning (no reward signal).}
\label{fig:robo-plan}
\vspace{-1em}
\end{figure}

\subsection{Results for Insertion and Manipulation Domains}
\vspace{-0.5em}
In practice, it might be very difficult to extract a context vector describing the environment every time the environment changes. In our third domain, we show that conditioning on a random image in that environmental configuration is sufficient in producing successful plans. Unlike previous block domains, we demonstrate the zero-shot generalization ability of our approach by varying the shape and volume of the moving object itself. The challenge in planning is accounting for the orientation of a novel shape when encountering obstacles, and figuring out the best angle at which to approach a narrow passageway. We emphasize that \textbf{such geometrical reasoning must be learned from the data, and must generalize to unseen shapes}.

For testing, we differentiated between `easy' tasks (ie. block stays on the same side of the wall) and `hard' tasks (ie. block must pass through the opening). Each task had 10 random start/goal locations, and all configurations were unseen. As seen in Figure \ref{fig:d3}, our method is successfully able to tackle all of these challenges. While successful on the majority of the `easy' tasks, Visual Foresight proved unable to plan the rotations necessary to move the block through the opening, and thus failed on most of the `hard' tasks. 

In addition, we applied HTM to robotic simulation of a Sawyer robot arm as seen in Figure \ref{fig:robo-plan} in which the robot needs to move a the green block to the desired location around a wall. We collect 45,000 samples of random interaction when the arm holds the green block, and 5,000 samples when the arm moves without the block. Here, we do not have different contexts, but we evaluate on unseen starts and goals. We apply HTM planning ability directly on real images from the replay buffer achieving feasible plans 12 out of 14 test tasks. We find that our visual plans avoid myopic behavior by planning to going around the thin wall, and preferring to grab the block before moving to the goal. 
% This domain is proposed as a simple yet challenging testbed for visual planning tasks.

%===============================================================================
\vspace{-0.5em}
\section{Discussion}
\vspace{-0.5em}
% We propose a method that is visually interpretable and modular -- we first hallucinate possible configurations, then compute a connectivity between them, and then plan. Our HTM can generalize to unseen environments and improve visual plan quality and execution success rate over state-of-the-art VP methods. 
We proposed a simple visual planning method that plans directly in image space, and generalizes in a zero-shot by hallucinating possible images conditioned on a domain context. On a suite of challenging visual planning domains, we find that our method outperforms state-of-the-art methods, and is able to pick up non-trivial geometrical information about objects in the image that is crucial for planning.
% is visually interpretable and modular -- we first hallucinate possible configurations, then compute a connectivity between them, and then plan. Our HTM can generalize to unseen environments and improve visual plan quality and execution success rate over state-of-the-art VP methods. 

Our results further suggest that combining classical planning methods with data-driven perception can be helpful for long-horizon visual planning problems, and takes another step in bridging the gap between learning and planning. In future work, we plan to combine HTM with Visual MPC for handling more complex objects, and use object-oriented planning for handling multiple objects. 
%Another interesting aspect is to improve planning by hallucinating samples conditioned on the start and goal configurations, which can help reduce the search space during planning.

\newpage
\bibliographystyle{icml2020}
\newpage

\bibliography{references}  % .bib
\newpage
\newpage
\appendix
\newpage
\clearpage

\section{Discriminative Models: Classifier vs. Energy model}
In this section, we assume the dataset as described in VPA, $\data = \{o^i_1, ..., o^i_{T_i}\}_{i=1}^n$. There are two ways of learning a model to distinguish the positive from the negative transitions.

\textbf{Classifier:} As noted above, SPTM first trains a classifier which distinguishes between an image pair that is within $h$ steps apart, and the images that are far apart using random sampling. The classifier is used to localize the current image and find possible next images for planning. In essence, the classifier contains the encoder $g_\theta$ that embeds the observation $x$ and the the score function $f$ that takes the embedding of each image and output the logit for a sigmoid function. The binary cross entropy loss of the classifier $L_{SPTM}(\theta, \psi; \data)$ is 
\vspace{-1em}
\begin{eqnarray*}
= -\sum_{(z_t, z_{t+k})\sim\data} (\log \frac{f_\psi(z, z_{t+k})}{1+f_\psi(z_t, z_{t+k})} \\
+ \log \frac{1}{1+f_\psi(z_t, z^{-}_{t})})  \\
= -\sum_{(z_t, z_{t+1})\sim \data} \log\left[\frac{f_\psi(z_t, z_{t+k})}{ f_\psi(z_t, z_{t+k}) + \alpha^t_\psi}\right]
\end{eqnarray*}
 where $\alpha^t_\psi$
= $1 + f_\psi(z_t, z^{-}_t) + f_\psi(z_t, z_{t+k})f_\psi(z_t, z_t^{-})$, and $z_t^{-}$ is a random  sample from $\data$.

\textbf{Energy model:} Another form of discriminating the the positive transition out of negative transitions is through an energy model. Oord et al. \citep{oord2018representation} learn the embeddings of the current states that are predictive of the future states. Let $g$ be an encoder of the input $x$ and $z = g_\theta(x)$ be the embedding. The loss function can be described as a cross entropy loss of predicting the correct sample from $N+1$ samples which contain $1$ positive sample and $N$ negative samples $L_{CPC}(\theta, \psi; \data)$ is 
\begin{align*}
= -\sum_{(z_t, z_{t+k})\sim \data}\log\left[\frac{f_\psi(z_t, z_{t+k})}{f_\psi(z_t, z_{t+k}) + \sum_{i=1}^N f_\psi(z_t, z^{i-}_t)}\right]
\end{align*}
where $f_\psi(u, v) = \exp{(u^T\psi v)}$ and $z^{1-}_t, ..., z^{N-}_t$ are the random samples from $\data$.

Note that when the number of negative samples is 1 the loss function resembles the SPTM.

\section{Mutual Information (MI)} 
This quantity measures how much knowing one variable reduces the uncertainty of the other variable. More precisely, the mutual information between two random variables $X$ and $Y$ can be described as 
\vspace{-1em}
\begin{align*}
 I(X, Y)= H(X) - H(X|Y) = H(Y) - H(Y|X) \\  = \mathbb{E}_{X, Y}\left[\frac{p_{X, Y}}{p_X p_Y}\right].  
\end{align*}
% \begin{proof}
% $I(X, Y) = H(X) - H(X|Y) = H(X) - H(X|Y, g(Y)) \geq H(X) - H(X|g(Y)) = I(X, g(Y))$
% \end{proof}
\vspace{-2em}

\section{Planning as Inference}

After training the CPC objective to convergence, we have $f_k(o_{t+k}, o_{t}) \propto p(o_{t+k}|o_t)/p(o_{t+k})$~\citep{oord2018representation}. To estimate $p(o_{t+k}|o_t)/p(o_{t+k}),$ we compute the normalizing factor $\sum_{o'\in V}f_k(o', o_t)$ for each $o_t$ by averaging over all nodes in the graph.
% \AT{are wesampling or averaging over neighbors in the graph?} \TK{The graph is fully connected}. 
Therefore, our non-negative weight from $o_t$ to $o_{t+k}$ is defined as $\omega(o_t, o_{t+k})=\sum_{o'\in V} f_k(o', o_t)/f_k(o_{t+k}, o_t) \approx p(o_{t+k})/p(o_{t+k}|o_t).$

A shortest-path planning algorithm finds $T, o_0, ..., o_T$ that minimizes $\sum_{t=0}^{T-1} \omega(o_t, o_{t+1})$ such that $o_0=o_{start}, o_T=o_{goal}$. By Jensen's inequality and the Markovian property of $o_0, ..., o_{T}$ we have that,
% \begin{align*}
$\log \frac{1}{T}\sum_{t=0}^{T-1} \omega(o_t, o_{t+1}) \geq \frac{1}{T}\sum_{t=0}^{T-1} \log \omega(o_t, o_{t+1})
= \frac{1}{T}\sum_{t=0}^{T-1} (\log p(o_{t+1}) - \log p(o_{t+1}|o_t))
= \frac{1}{T}\sum_{t=1}^{T-1}p(o_t) - \log p(o_1, ..., o_{T-1}|o_0 = o_{start}, o_T=o_{goal}).$
% \end{align*}
% \begin{align*}
% \log \frac{1}{T}\sum_{t=0}^{T-1} \omega(o_t, o_{t+1}) &\geq \frac{1}{T}\sum_{t=0}^{T-1} \log \omega(o_t, o_{t+1}) \\
% &= \frac{1}{T}\sum_{t=0}^{T-1} (\log p(o_{t+1}) - \log p(o_{t+1}|o_t)) \\
% &= \frac{1}{T}\sum_{t=1}^{T-1}p(o_t) - \log p(o_1, ..., o_{T-1}|o_0 = o_{start}, o_T=o_{goal}),
% \end{align*}
% Thus, among all paths of length $T$ from start $s$ to goal $g_{enc}$, the maximum score path based on graph weight $\log f_k(x_{t+k}, x_t)$ corresponds to 
% \begin{align*}
%     \max_{\substack{x_1, ... x_{T-1}\\x_0=s, x_T=g}} \sum_{t=0}^{T} \log f_k(x_{t+k}, x_t) = \max_{\substack{x_1, ... x_{T-1}\\x_0=s, x_T=g}} \sum_{t=0}^{T}  \log p(x_{t+k}|x_t) - \log p(x_t)
% \end{align*}
% Because the sequence $X_0, ..., X_T$ is a markov chain, the objective becomes
% \[
% \max_{x_1, ... x_{T-1}} \log p(x_0=s, x_1, ..., x_{T-1},x_T=g) - \sum_{t=1}^{T-1} \log p(x_t)
% \]
% achieved through a Markovian property of $o_0, ..., o_{T}$. 
Thus, since $p(o_t)$ is fixed by uniform asssumption, the shortest path algorithm with proposed weight $\omega$ maximizes a lower bound on the trajectory likelihood given the start and goal states. In practice, this leads to a more stable planning approach and yields more feasible plans.

\section{Block Insertion Domain} 
\begin{figure}[h]
\centering
\includegraphics[width=.3\textwidth]{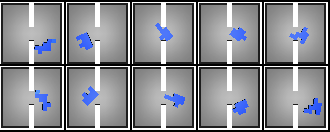}
\caption{Example of observations (top) and contexts (bottom) of block insertion domain. } 
\label{fig:d3ex}
\end{figure}

In this domain, we kept the obstacle constant and varied the agent itself. In particular, we uniformly chose from 4 to 10 units, with 6 as the holdout, and then randomly placed those units such that they resembled a contiguous shape. When applying an action, we applied a vertical and horizontal force to the middle block, and also a rotation force on the first and last unit laid down, leading to a total action space of four. As our context vector, we randomly chose any image from all trajectories with that same context, as seen in Figure \ref{fig:d3ex}. During testing time, we randomly generated shapes from 3, 6, and 11 units. The L2 threshold distance for success was thus the total L2 distance for all units divided by the number of units.

\section{Additional Results and Hyperparameters} \label{sec: hp}
\vspace{-0.5em}

% HTM plans
\vspace{-1em}
\begin{figure*}[h]
\centering
\includegraphics[width=0.8\textwidth]{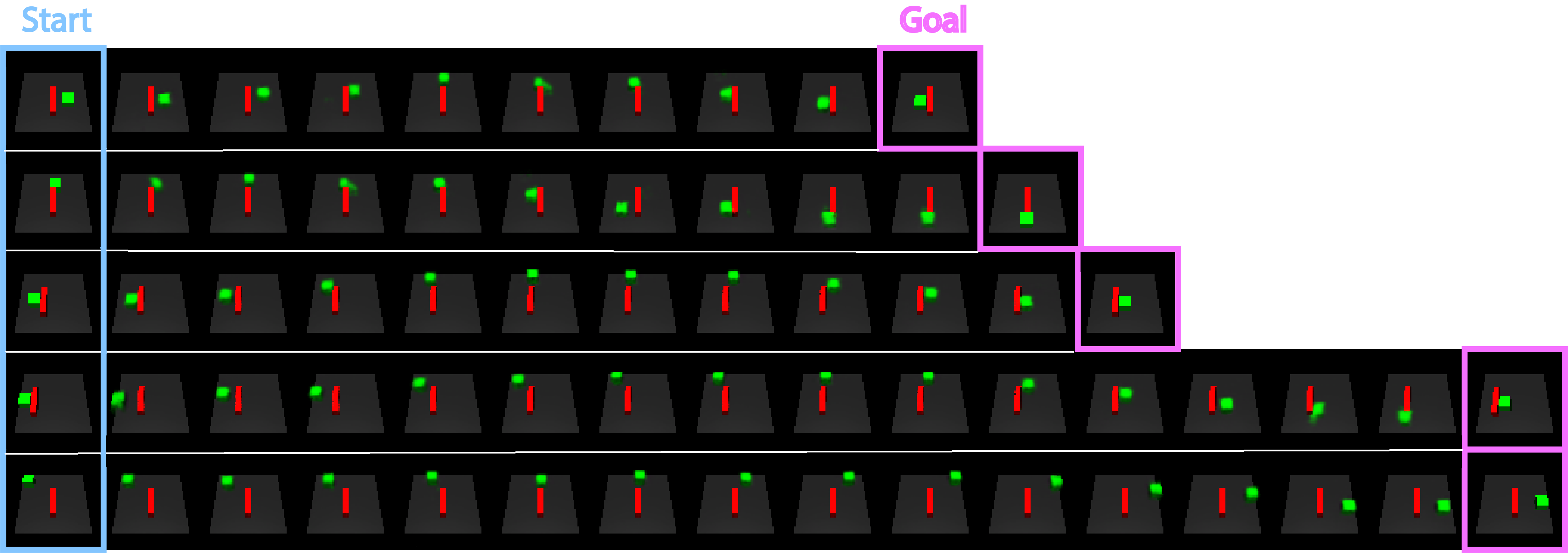}
\caption{HTM plan examples on the block wall domain. The hallucination allows the planner to imagine how to go around the wall even though it has not seen the context before.}
\end{figure*}
\vspace{-2em}

% Visual Foresight
\begin{figure*}[h]
\centering
\includegraphics[width=.8\textwidth]{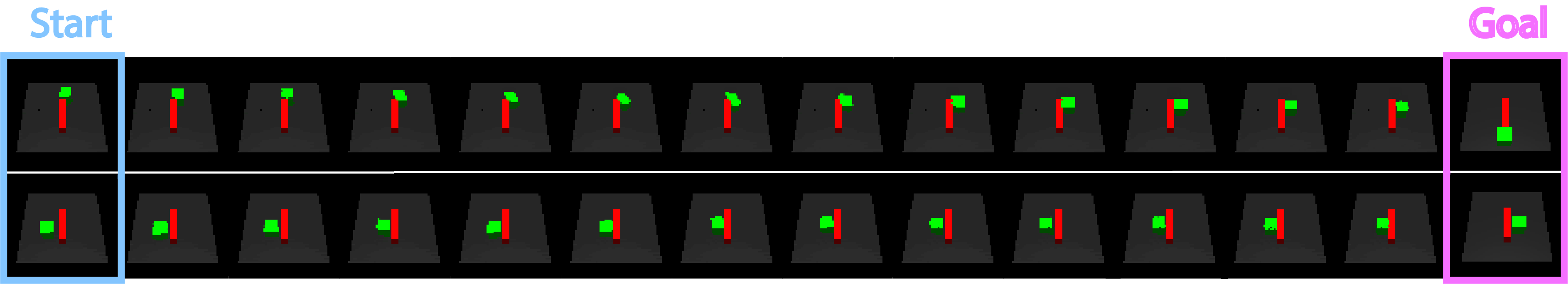}
\caption{Visual Foresight plan examples on the block wall domain. The plans do not completely show the trajectory to the goal.}
\label{fig:vf_ex}
\end{figure*}
\vspace{-2em}

\begin{table*}[h!]
\centering
\caption{Data parameters.} 
\begin{tabular}{c|c|c|c|c } 
& Domain 1 & Domain 2 & Domain 3 & Domain 4 \\
 \hline 
 no. contexts & 150 & 400 & 360 & 1\\
 initializations per context & 50 & 30 & 20 & 1000\\ 
trajectory length & 20 & 100 & 50 & 50\\
action space & $ [-.05,.05]^2 $ & $[-.1,.1]^2$ & $[-.05,.05]^4$ & $[-1, 1]^2$\\ 
table size & 2.8x2.8 & 2.8x2.8 & .8x.8 & .9x.7
\end{tabular}
\vspace{1ex}
\end{table*}
\vspace{-2em}

\begin{table*}[h!]
\centering
\caption{Planning hyperparameters.}
\begin{tabular}{c|c|c |c} 
& Domain 1 & Domain 2 & Domain 3 \\
 \hline 
 no. of samples from CVAE & 300 & 500 & 300  \\
 L2 threshold for success (for each unit) & .5 & .75 & .1  \\ 
 $n$ (timesteps to get to goal) & 500 & 400 & 400  \\
 $r$ (timesteps until replanning) & 200 & 80 & 80  \\ 
\end{tabular}
\vspace{1ex}
\end{table*}
\vspace{-2em}

\end{document}